\documentclass[10pt,twocolumn,letterpaper]{article}

\usepackage{cvpr}              
\usepackage{graphicx}
\usepackage{amsmath}
\usepackage{amssymb}
\usepackage{booktabs}
\usepackage{dsfont}
\usepackage[mathscr]{euscript}

\newcommand{\ra}[1]{\renewcommand{\arraystretch}{#1}} 
\usepackage[table]{xcolor}
\usepackage{color, colortbl,xcolor}
\definecolor{Gray}{gray}{0.9}
\definecolor{light-gray}{gray}{0.95}

\usepackage{pifont}

\usepackage{booktabs} 
\usepackage{multirow} 
\usepackage{subcaption} 
\usepackage{url} 

\usepackage{cuted} 
\usepackage{capt-of}
\usepackage[font={small}]{caption}
\usepackage{epigraph}
\usepackage{comment}
\usepackage{enumitem}

\providecommand\para[1]{\medskip \noindent \textbf{#1}}

\usepackage[pagebackref,breaklinks,colorlinks]{hyperref}

\usepackage[capitalize]{cleveref}
\crefname{section}{Sec.}{Secs.}
\Crefname{section}{Section}{Sections}
\Crefname{table}{Table}{Tables}
\crefname{table}{Tab.}{Tabs.}

\def\cvprPaperID{2882} 
\def\confName{CVPR}
\def\confYear{2023}

\begin{document}

\title{From Audio to Photoreal Embodiment: Synthesizing Humans in Conversations}


\author{Evonne Ng\textsuperscript{1, 2}
\hspace{0.3in} Javier Romero\textsuperscript{1}
\hspace{0.3in} Timur Bagautdinov\textsuperscript{1}
\hspace{0.3in} Shaojie Bai\textsuperscript{1} \\
\hspace{0.3in} Trevor Darrell\textsuperscript{2} 
\hspace{0.3in} Angjoo Kanazawa\textsuperscript{2}
\hspace{0.3in} Alexander Richard\textsuperscript{1}
\vspace{10pt}
\\
\textsuperscript{1}{Codec Avatars Lab, Meta, Pittsburgh}
\hspace{0.3in} \textsuperscript{2}{University of California, Berkeley} \\ 
}
\maketitle


\begin{strip}\centering
\includegraphics[width=\linewidth]{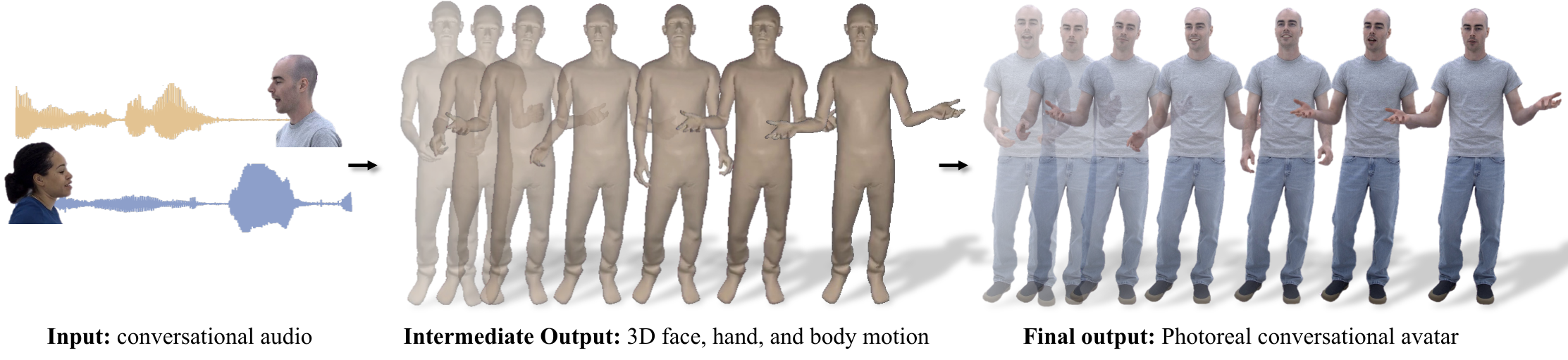}
\vspace{-0.5cm}
\captionof{figure}{{\bf Synthesizing photoreal conversational avatars.} Given the audio from a dyadic conversation, we generate realistic conversational motion for the face, body, and hands. The motion can then be rendered as a photorealistic video. Please see \href{https://youtu.be/Y0GMaMtUynQ}{results video}.
}
\label{fig:teaser}
\end{strip}


\begin{abstract}
We present a framework for generating full-bodied photorealistic avatars that gesture according to the conversational dynamics of a dyadic interaction.
Given speech audio, we output multiple possibilities of gestural motion for an individual, including face, body, and hands.
The key behind our method is in combining the benefits of sample diversity from vector quantization with the high-frequency details obtained through diffusion to generate more dynamic, expressive motion.
We visualize the generated motion using highly photorealistic avatars that can express crucial nuances in gestures (\eg sneers and smirks).
To facilitate this line of research, we introduce a first-of-its-kind multi-view conversational dataset that allows for photorealistic reconstruction. 
Experiments show our model generates appropriate and diverse gestures, outperforming both diffusion- and VQ-only methods. Furthermore, our perceptual evaluation highlights the importance of photorealism (vs.~meshes) in accurately assessing subtle motion details in conversational gestures.
Code and dataset available on \href{https://people.eecs.berkeley.edu/~evonne_ng/projects/audio2photoreal/}{project page}.
\end{abstract}

\section{Introduction}

Consider talking to your friend in a telepresent world, where they appear as the generic golden mannequin shown in Figure~\ref{fig:teaser} (middle).
Despite the mannequin's ability to 
act out rhythmic strokes of arm motion that seemingly 
follow your friend's voice,
the interaction will inevitably feel robotic and uncanny.
This uncanniness stems from the limitations imposed by non-textured meshes which mask subtle nuances like eye gaze or smirking. 
Photorealistic details can effectively convey these nuances, allowing us to express diverse moods during conversation. For example, a sentence spoken while avoiding eye contact differs significantly from one expressed with sustained gaze.
As humans, we are especially perceptive to these micro-expressions and movements, which we use to formulate a higher-order understanding of our conversational partner's intentions, comfort, or understanding~\cite{ekman1969nonverbal}.
Developing conversational avatars with the level of photorealism that can capture these subtleties is therefore essential for virtual agents to meaningfully interact with humans.

Our ability to perceive these fine-grain motion patterns breaks down as we represent the motion in more abstracted forms.
Chaminade \etal~\cite{chaminade2007anthropomorphism} demonstrates that humans have a more difficult time distinguishing real vs.~fake key-framed motions (such as walking) in skeletons than in textured meshes, and even more-so in point-based representations than in skeletons. In faces, McDonnell \etal~\cite{mcdonnell2012render} shows that large facial motion anomalies are considerably less discernible on Toon (i.e. plain colored, comic-like) characters, than on characters with human textures applied. 
Although abstract representations cannot precisely represent the level of detail needed for humans to interpret subtle conversational cues, 
the majority of prior works in gesture generation~\cite{yi2023generating, alexanderson2023listen, joo2019towards, lee2019talking} still assess their methods using mesh-based or skeletal representations.
In this paper we advocate the importance of developing \emph{photorealistic} conversational avatars which not only allow us to express subtle motion patterns, but also allow us to more accurately evaluate the realism of the synthesized motion.

To this end, we present a method for generating photorealistic avatars, conditioned on the speech audio of a dyadic conversation.
Our approach synthesizes diverse high-frequency gestures (e.g.~pointing and smirking)
 and expressive facial movements that are well-synchronized with speech.
For the body and hands, we leverage advantages of both an autoregressive VQ-based method and a diffusion model.
Our VQ transformer takes conversational audio as input and outputs a sequence of guide poses at a reduced frame rate, allowing us to sample diverse poses (e.g.~pointing) while avoiding drift. We then pass both the audio and guide poses into the diffusion model, which infills intricate motion details (e.g.~finger wag) at a higher fps.
For the face, we use an audio conditioned diffusion model. 
The predicted face, body, and hand motion are then rendered with a photorealistic avatar.
We demonstrate the added guide pose conditioning on the diffusion model allows us to generate more diverse and plausible conversational gestures compared to prior works.
In a perceptual study, we further illustrate that evaluators can better distinguish differences between two approaches when motion is visualized with photorealistic avatars than with meshes.

To support our approach in modeling the intricacies of human conversation, we introduce a rich dataset of dyadic interactions captured in a multi-view system. This system allows for highly accurate body/face tracking and photorealistic 3D reconstructions of both participants simultaneously.  
The non-scripted, long-form conversations cover a wide range of topics and emotions.
In contrast to prior full-body datasets that support skeletal~\cite{joo2019towards, lee2019talking} or Toon-like visualizations~\cite{liu2022beat}, we reconstruct photorealistic renders of each individual in the dataset. Our data also captures the dynamics of inter-personal conversations rather than individual monologues~\cite{ginosar2019learning, liu2022beat, yi2023generating}.
We will release the dataset and renderer, and hope these will encourage the investigation of gesture generation in a photorealistic manner.

To the best of our knowledge, we are the first to investigate the generation of photorealistic face, body, \emph{and} hand motion for interpersonal conversational gestures. Our VQ- and diffusion-based method synthesizes more realistic and diverse motion compared to prior works. 
Furthermore, we pose an important question on the validity of evaluating conversational motion using non-textured meshes, as humans may overlook or be less critical of inaccuracies in these representations.
Finally, to support this investigation, we introduce a novel dataset of long-form conversations that enable renderings of photorealistic conversational avatars. Code, dataset, and renderers will all be publicly available.


\section{Related Work}
\label{sec:relatedworks}

\begin{figure}[t]
    \centering
    \includegraphics[width=\linewidth]{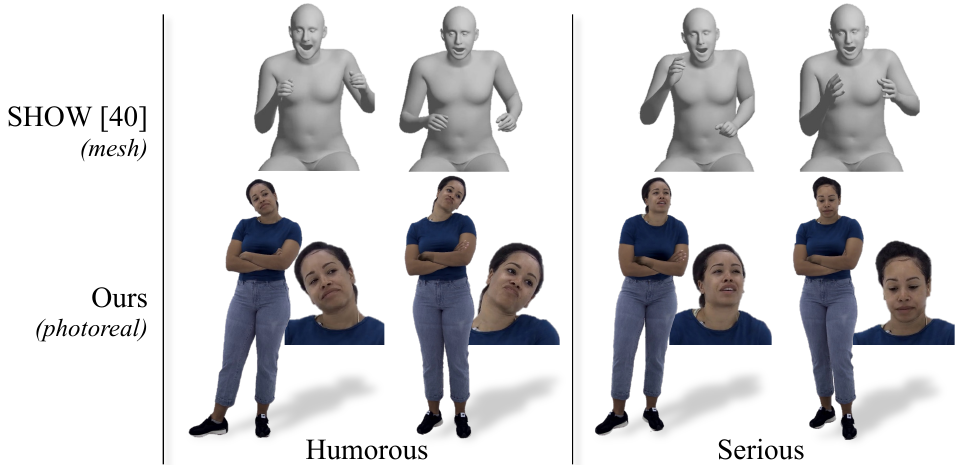}
    \caption{\textbf{Importance of photorealism} 
    Top: Mesh annotations from prior work~\cite{yi2023generating}. Bottom: Our photorealistic renderings.
    For the mesh, differences in laughing (top left) vs.~speaking (top right) are difficult to perceive. 
    In contrast, photorealism allows us to capture subtle details such as the smirk (bottom left) vs.~grimace (bottom right), which completely changes the perception of her current mood despite similar coarse body poses.
    }
    \label{fig:realism}
\end{figure}

\noindent\textbf{Interpersonal conversational dynamics.}
Traditionally, animating conversational avatars have involved constructing rule-based guides from lab captured motion data \cite{Cassell1994, gratch2006virtual, huang2011virtual, Bohus_Horvitz_2010}. 
These methods are often limited in variety of gestures and rely on simplifying assumptions that do not hold on in-the-wild data. 
As a result, there has been greater focus on using learning-based methods to predict coarse aspects of a conversation such as turn-taking~\cite{joo2019towards, ahuja2019react} or a single facial expression to summarize a conversation~\cite{huang2017dyadgan,nojavanasghari2018interactive}. 
While these methods focus on higher-level dynamics, our method focuses on the lower-level complexities of interactions by modeling the full range of facial expressions and body-hand motion. In contrast, Tanke~\etal~\cite{tanke2023social} predicts the full body pose, but focuses on a different task of motion forecasting, where the goal is to generate plausible future body poses for a triad given their past body motion.

More recently, there have been works on modeling cross-person interaction dynamics by predicting the listener’s fine-grain 2D \cite{learn2smile2017} or 3D gestural motion from the speaker’s motion and audio \cite{jonell2019learning,ng2022learning}, text \cite{Ng_2023_ICCV}, or stylized emotion \cite{zhou2022responsive}. 
However, all these methods generate only the head pose and facial expression of the listener alone.
On the other extreme, Lee~\etal~\cite{lee2019talking} models only the finger motion of the speaker in a dyadic conversation.
In contrast, our method is the first to consider the full range of 3D face, body, and hand motion for interpersonal conversation while using a single model to handle both speaking and listening motion.

\begin{figure}[t]
    \centering
    \includegraphics[width=\linewidth]{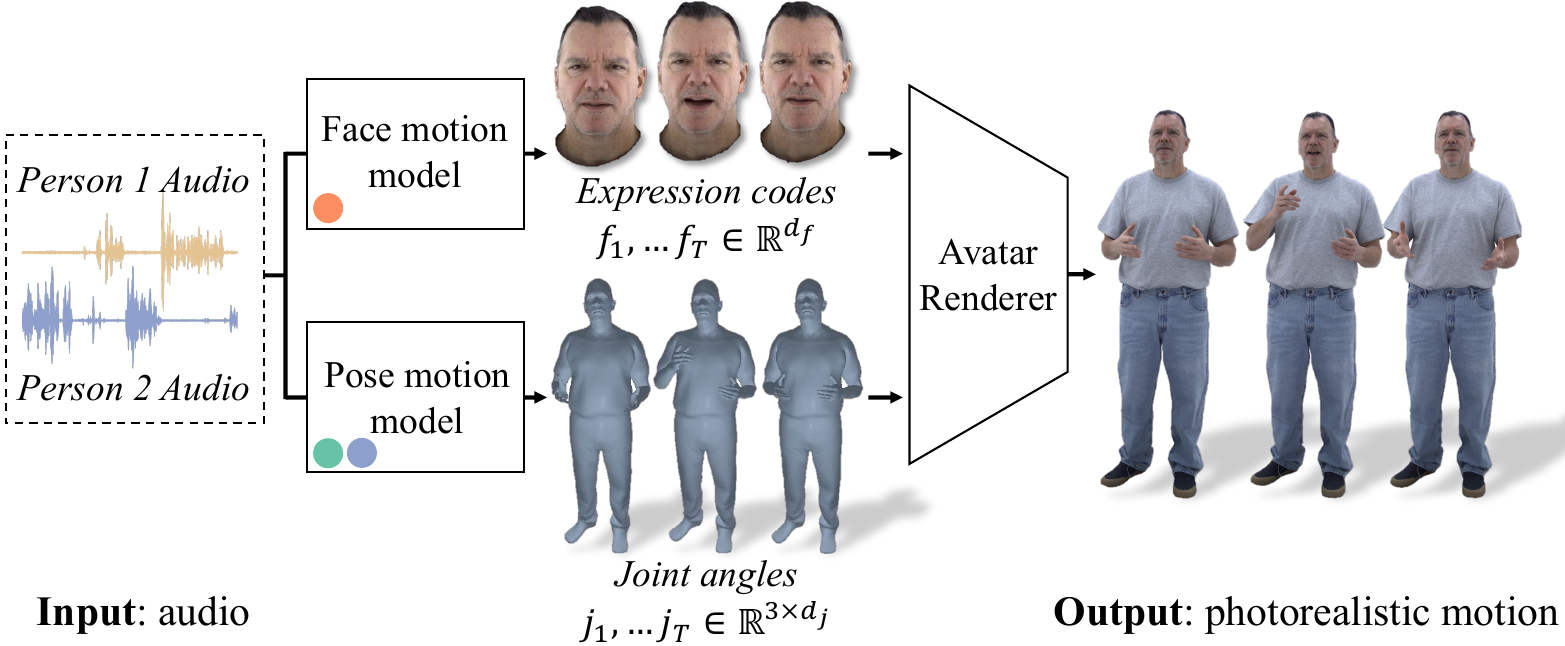}
    \caption{\textbf{Method Overview} Our method takes as input conversational audio and generates corresponding face codes and body-hand poses. The output motion is then fed into our trained avatar renderer, which generates a photorealistic video. For details on the face/pose models, please see Figure~\ref{fig:method}.}
    \label{fig:method_overview}
\end{figure}

\begin{figure*}[t]
    \centering
    \includegraphics[width=\textwidth]{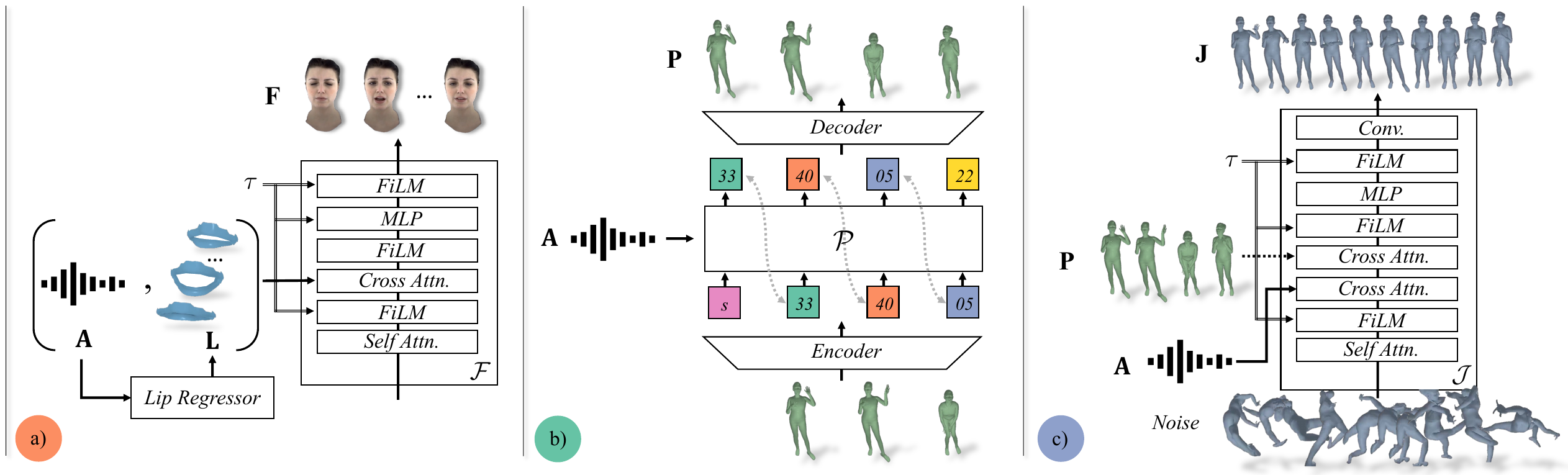}
    \vspace{-0.5cm}
    \caption{\textbf{Motion generation} (a) Given conversational audio $\mathbf{A}$, 
    we generate facial motion $\mathbf{F}$ using a diffusion network conditioned on both audio and the output of a lip regression network $\mathbf{L}$, which predicts synced lip geometry from speech audio.
    (b) For the body-hand poses, we first autoregressively generate guide poses $\mathbf{P}$ at a low fps using a VQ-Transformer. (c) The pose diffusion model then uses these guide poses and audio to produce a high-frequency motion sequence $\mathbf{J}$.
    }
    \label{fig:method}
    \vspace{-0.2cm}
\end{figure*}

\para{Gestural motion generation.}
Prior works on diffusion have explored
audio to dance~\cite{tseng2023edge}, text to motion~\cite{tevet2022human}, or even audio to gestures~\cite{alexanderson2023listen, ao2023gesturediffuclip, yu2023talking, Zhi_2023_ICCV}. In
~\cite{alexanderson2023listen, ao2023gesturediffuclip}, body motion of a speaker is synthesized using a diffusion model conditioned on audio or text respectively. 
Meanwhile, Yu~\etal~\cite{yu2023talking} focuses only on the face by using a diffusion-based method with contrastive learning to produce lip sync that is both accurate and can be disentangled from lip-irrelevant facial motion.
While these methods model only the body or the face, our approach generates the full \emph{face, body, and hands} of the conversational agent simultaneously.

SHOW~\cite{yi2023generating} addresses this issue by training separate VQ’s to produce face, body, and hand motion given a speaker's audio.
While our approach similarly focuses on generating the full range of face, body, and hand motion for a conversational agent, our approach significantly differs 
in that we visualize on photorealistic avatars as opposed to mesh-based renderings. As depicted in Figure~\ref{fig:realism}, their mesh can represent large arm movements that follow a rhythm, but struggles to capture crucial distinctions between a laugh and opening one's mouth to speak (top). In contrast, we are the first to employ photoreal avatars that can express subtle nuances such as a grimace vs.~a smirk (bottom). 
We demonstrate in our analysis (Sec.~\ref{sec:perceptual}) that photorealism greatly affects the evaluation paradigm for conversational agents.

We further differentiate from these prior works~\cite{alexanderson2023listen, ao2023gesturediffuclip, yu2023talking, yi2023generating} in that we model \emph{interpersonal} communication dynamics 
of a dyadic conversation as opposed to a single speaker in a monadic setting. 
As a result, our method must model both listener and speaker motion, and generate motion that not only looks realistic with respect to the audio, but also reacts realistically to the other individual in conversation. 

\para{Conversational datasets.}
There is a growing number of large scale datasets for conversational motion
~\cite{liu2022beat, yi2023generating, lee2019talking, ng2021body2hands}. Pose parameters for the face, body and hands of a monologue speaker are released at large scale in ~\cite{liu2022beat, yi2023generating}. Similarly~\cite{lee2019talking, ng2022learning} provide only the body and hand reconstructions. However, all these datasets release only enough information to reconstruct coarse human meshes or textured avatars through blendshapes that lack photorealism and high-frequency details~\cite{liu2022beat}.

Given the popularity of the task of audio-driven lip syncing, there are many datasets with open-sourced pipelines for generating facial motion~\cite{yu2023talking, ji2021audio, thies2020neural, vougioukas2020realistic, zhang2020apb2face, richard2021meshtalk, cudeiro2019voca}, though these approaches are limited to either 2D video or 3D mesh-based animation. Complementing such work with a focus on the face, Ginosar~\etal~\cite{ginosar2019learning} provides a way to render out the body and hands of a monologue speaker. To the best of our knowledge, we are the first to provide a dataset with full simultaneous reconstructions of the~\emph{face, body, and hands}, and to consider this in a~\emph{dyadic} conversational setting.

\section{Photoreal full body motion synthesis}
\label{sec:method}

Given raw audio from a conversation between two people, we introduce a model that generates corresponding photorealistic face, body, and hand motion for one of the agents in the dyad.
We represent the face as latent expression codes from the recorded multi-view data following~\cite{lombardi2018deep}, and the body pose as joint angles in a kinematic skeleton.
As shown in Fig.~\ref{fig:method_overview}, our system consists of two generative models that produce sequences of expression codes and body poses given audio from the dyadic conversation as input.
Expression codes and body pose sequences can then be rendered frame-by-frame using our trained neural avatar renderer~\cite{bagautdinov2021bodies} which produces the full textured avatar with the face, body, and hands from
a given camera view.\footnote{Face expression codes, tracked joint angles, and the pre-trained full-body renderer are released as part of the dataset.}

Note that the body and face follow highly different dynamics.
First, the face is strongly correlated with the input audio, particularly in terms of lip motion, while the body has a weaker correlation with speech. This leads to greater diversity in plausible body gestures for a given speech input.
Second, since we represent face and body in two different spaces (learned expression codes vs.\ joint angles), each of them follow different temporal dynamics.
We therefore model the face and body with two separate motion models. This allows the face model to spend its capacity on generating speech-consistent facial details, and the body model to focus on generating diverse yet plausible body motion. 

The \textbf{face motion model} is a diffusion model conditioned on input audio and lip vertices produced by a pre-trained lip regressor (Fig.~\ref{fig:method}a).
For the \textbf{body motion model}, we found that a purely diffusion-based model conditioned only on audio produces less diverse motion that appears temporally uncanny.
However, the quality improves when we condition on diverse guide poses. 
We therefore split the body motion model into two parts:
First, an autoregressive audio-conditioned transformer predicts coarse guide poses at 1fps (Fig.~\ref{fig:method}b), which are then consumed by the diffusion model to in-fill fine grain and high-frequency motion (Fig.~\ref{fig:method}c).
We describe the model components in detail below.

\para{Notation.}
We denote the audio of the agent as $ \mathbf{a}_{\mathit{self}} $ and the audio of the conversation partner as $ \mathbf{a}_{\mathit{other}} $.
For both audio streams, we extract Wav2Vec~\cite{baevski2020wav2vec} features such that the audio input is $ \mathbf{A} = (\mathbf{a}_{\mathit{self}}, \mathbf{a}_{\mathit{other}}) \in \mathds{R}^{2 \times d_a \times T} $, with $ d_a $ denoting the feature dimension of Wav2Vec features. 

We denote a sequence of $T$ face expression codes as $ \mathbf{F} = (f_1,\dots,f_T) $, where each $ f_t \in \mathds{R}^{256} $ represents a face expression for frame $t$.
A body motion sequence of $T$ frames is represented by $ \mathbf{J} = (j_1,\dots,j_T) $, where $ j_t \in \mathds{R}^{d_j \times 3} $ is a vector containing three rotation angles for each of the $ d_j $ body joints that define a pose at frame $t$.
We follow the forward kinematic representation~\cite{bagautdinov2021bodies}, where the body and hand pose of a given person can be constructed from the relative rotations of each joint with respect to its parent joint.

\begin{figure*}[t]
    \centering
    \includegraphics[width=\linewidth]{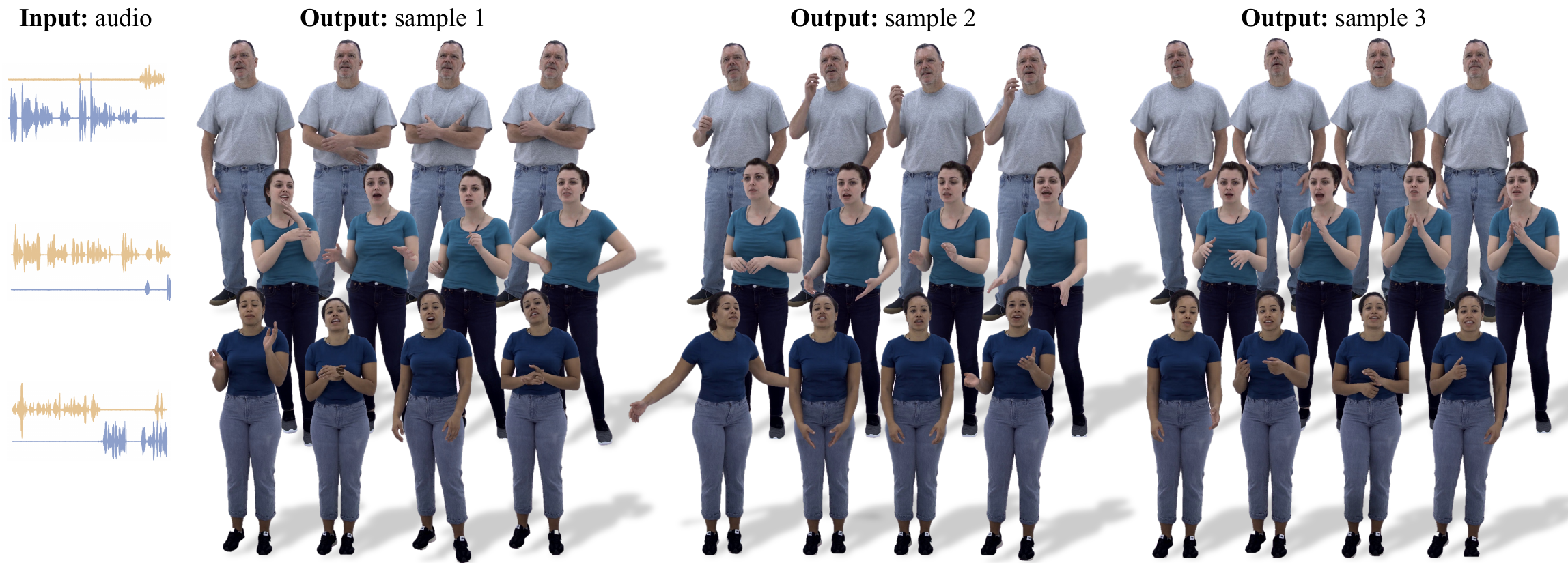}
    \vspace{-0.5cm}
    \caption{\textbf{Diversity of guide pose rollouts} Given the input audio for the conversation (predicted person's audio in gold), the transformer $\mathcal{P}$ generates diverse samples of guide pose sequences with variations in listening reactions (top), speech gestures (middle), and interjections (bottom). Sampling from a rich codebook of learned poses, $\mathcal{P}$ can produce ``extreme" poses e.g.~ pointing, itching, clapping, etc. with high diversity across different samples. These diverse poses are then used to condition the body diffusion model $\mathcal{J}$.}
    \label{fig:keyframes}
    \vspace{-0.25cm}
\end{figure*}

\subsection{Face Motion Diffusion Model}
\label{sec:fine}

To generate facial motion from audio input, we construct an audio-conditioned diffusion model.
We follow the DDPM~\cite{ho2020denoising} definition of diffusion. The forward noising process is defined as:
\begin{equation}
    q(\mathbf{F}^{(\tau)} | \mathbf{F}^{(\tau-1)}) \sim \mathcal{N}\big(\sqrt{\alpha}_\tau \mathbf{F}^{(\tau-1)}, (1 - \alpha_\tau) \mathbf{I}\big),
\end{equation}
where $ \mathbf{F}^{(0)}$ approximates the clean (noise-free) sequence of face expression codes $\mathbf{F}$, 
$ \tau \in [1,\dots,\dot{T}] $ denotes the forward diffusion step, and $ \alpha_\tau \in (0,1)$ follows a monotonically decreasing noise schedule such that as $\tau$ approaches $\dot{T}$, we can sample 
$\mathbf{F}^{(\dot{T})} \sim \mathcal{N}(0, \mathbf{I})$.

To reverse the noising process, we follow~\cite{ho2020denoising,nichol2021improved} and define a model to denoise $ \mathbf{F}^{(0)} $ from the noisy $ \mathbf{F}^{(\tau)} $. 
The next step $ \mathbf{F}^{(\tau-1)} $ of the reverse process can then be obtained by applying the forward process to the predicted $ \mathbf{F}^{(0)} $.
We predict $ \mathbf{F}^{(0)} $ with a neural network $ \mathcal{F} $:
\begin{equation}
    \mathbf{F}^{(0)} \approx \mathcal{F}(\mathbf{F^{(\tau)}}; \tau, \mathbf{A}, \mathbf{L}),
\end{equation}
where $ \mathbf{A} $ are the input audio features and $ \mathbf{L} = (l_1,\dots,l_T) $ is the output of a pre-trained audio-to-lip regressor following~\cite{cudeiro2019voca}, but limited to lip vertices instead of full face meshes.
We train the lip-regressor on $30$h of in-house 3D mesh data.
Each $ l_t \in \mathds{R}^{d_l \times 3} $ is a predicted set of $d_l$ lip vertices at frame $t$ given audio $\mathbf{A}$. Tab.~\ref{tab:lip} shows, conditioning on both the lip regressor output and audio significantly improves lip sync quality over conditioning on audio alone.

The diffusion model is trained with the simplified ELBO objective~\cite{ho2020denoising},
\begin{equation}
    \mathcal{L}_{\mathit{simple}} = \mathds{E}_{\tau, \mathbf{F}}\big[\mathbf{F} - \mathcal{F}(\mathbf{F^{(\tau)}}; \tau, \mathbf{A}, \mathbf{L})\big].
\end{equation}

We train our model for classifier-free guidance~\cite{ho2021classifier} by randomly replacing either conditioning with $\mathbf{A}=\emptyset$ and $\mathbf{L}=\emptyset$ during training with low probabilities. To incorporate the audio and lip vertex information, we use a cross attention layer. Timestep information is incorporated with a feature-wise linear modulation (FiLM) layer, see Fig.~\ref{fig:method}a.


\subsection{Body Motion Model}
\label{sec:coarse}


To generate body motion, we extend the conditional diffusion model by introducing guide poses sampled at 1fps as additional conditioning. This allows us to model more expressive motion.
Similar to the face model that did not generate accurate lip motion when conditioned on audio alone, we found that the body model generates less plausible motion with limited diversity when conditioned on audio only.

More formally, to generate a full body motion sequence at 30fps, we train the body diffusion model with guide poses $ \mathbf{P} = \{j_{k \cdot 30} | 1 \leq k \leq T / 30\} $ taken at 1fps.
These guide poses are obtained by subsampling the original 30 fps ground truth body pose sequence $ \mathbf{J} $.
The body motion diffusion model $ \mathcal{J} $ is then the same network as the face motion diffusion model $ \mathcal{F} $, but is conditioned on the subsampled guide poses, \ie $ \mathbf{J}^{(0)} \approx \mathcal{J}(\mathbf{J^{(\tau)}}; \tau, \mathbf{A}, \mathbf{P}) $.
The guide poses are incorporated using an additional cross attention layer (see Fig.~\ref{fig:method}c). 
At inference time, however, ground truth guide poses are not available and need to be generated.

\para{Guide pose generation.}
To generate guide-poses at inference time, we train an autoregressive transformer to output coarse keyframes at 1fps that adhere to the conversational dynamics.
As autoregressive transformers typically operate on discrete tokens~\cite{esser2021taming, ng2022learning, yi2023generating}, we first quantize the 1 fps guide pose sequence using a residual VQ-VAE~\cite{vasuki2006review}.
Residual VQ-VAEs are similar to vanilla VQ-VAEs~\cite{van2017neural}, but they recursively quantize the residuals of the previous quantization step instead of stopping after a single quantization step. This leads to higher reconstruction quality~\cite{vasuki2006review,zeghidour2021soundstream}.

Let $ \mathbf{Z} = (z_1,\dots,z_K) $ be the resulting quantized embedding of the $K$-length guide pose sequence $ \mathbf{P} $, where $ z_k \in \{1,\dots,C\}^N $, $ C $ is codebook size, and $ N $ is the number of residual quantization steps.
We flatten this $ K \times N $-dimensional quantized embedding $ \mathbf{Z} $ to obtain $ \mathbf{\hat Z} = (\hat z_{1},\dots,\hat z_{K \cdot N}) $. We predict $ \mathbf{\hat Z} $ with an audio-conditioned transformer $ \mathcal{P} $, which outputs a categorical distribution over the next token given prior predictions and audio,
\begin{equation}
    p(\hat z_k | \hat z_{1:k-1}, \mathbf{A}) = \mathcal{P}(\hat z_{1:k-1}; \mathbf{A}).
\end{equation}
We train the transformer using a simple cross entropy loss on the task of next-token prediction with teacher forcing:
\begin{equation}
    \mathcal{L}_{\mathcal{P}} = -\sum_{k \in \{1, \dots, K \cdot N\}} \log\Pr\left[\mathcal{P}(z_{1:k-1}, \mathbf{A}) = z_k\right].
\end{equation}
At test time, we use nucleus sampling~\cite{holtzman2019curious} to predict the sequence of motion tokens. We can easily control the level of variability seen across samples by increasing or decreasing the cumulative probability. 

The guide-pose transformer is illustrated in Fig.~\ref{fig:method}b. For further architecture details on the residual VQ-VAE and the transformer architecture refer to Appendix~\ref{app:method}.

\begin{figure*}[t]
    \centering
    \includegraphics[width=\linewidth]{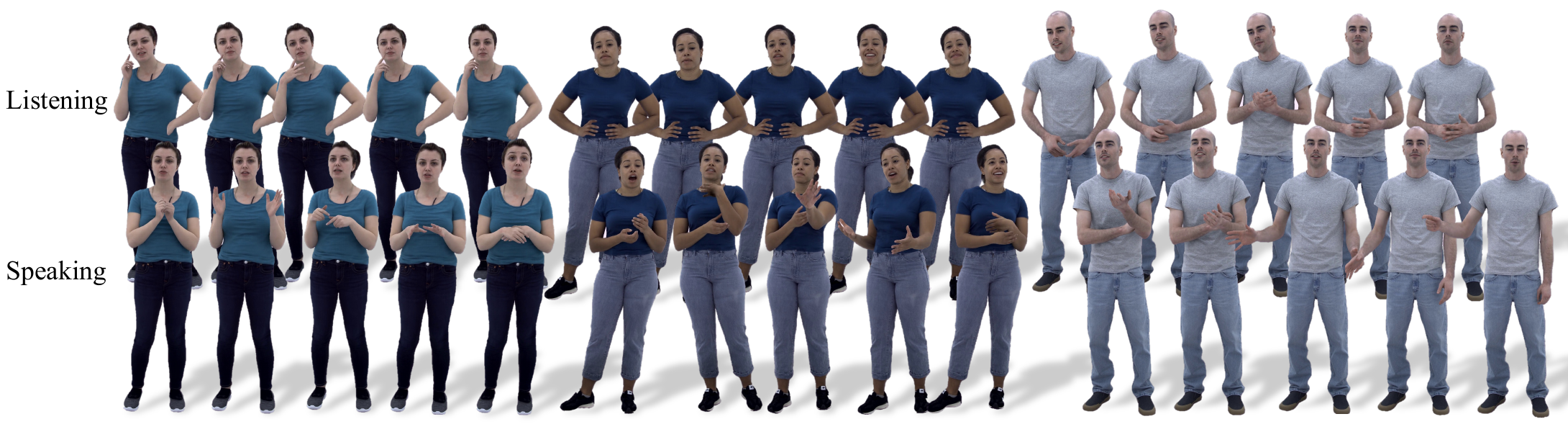}
    \vspace{-0.6cm}
    \caption{\textbf{Results} Our method produces gestural motion that is synchronous with the conversational audio. During periods where the person is listening (top), our model correctly produces still motion, seemingly as if the avatar is paying attention. In contrast, during periods of talking (bottom), the model produces diverse gestures that move synchronously with the audio.} 
    \label{fig:analysis}
    \vspace{-0.3cm}
\end{figure*}

\subsection{Photorealistic Avatar Rendering}

Given both the generated facial expression sequence $ \mathbf{F} $ and the generated body pose sequence $ \mathbf{J} $, the full photorealistic avatar can be rendered as illustrated in Fig.~\ref{fig:method_overview}. 
Following~\cite{bagautdinov2021bodies}, we use a learning-based method to build our drivable avatars.
The model takes as input one frame of facial expression $f_t$, one frame of body pose $j_t$, 
and a view direction. It then outputs a registered geometry and view-dependent texture, which is used to synthesize images via rasterization.
The model is a conditional variational auto-encoder (cVAE) consisting of an encoder and decoder, both parameterized as convolutional neural networks. 
The cVAE is trained end-to-end in a supervised manner to reconstruct images of a subject captured in a multi-view capture setup.
We train a personalized avatar renderer for each subject in our dataset. For details, please refer to ~\cite{bagautdinov2021bodies}.


\section{Photorealistic conversational dataset}
\label{sec:dataset}

While there are a plethora of datasets on dyadic interactions~\cite{ng2022learning, lee2019talking}, all such datasets are only limited to upper body or facial motion.
Most related is Joo~\etal~\cite{joo2019towards}, which introduces a small-scale dataset of triadic interactions as a subset of the Panoptic Studio dataset. The data includes 3D skeletal reconstructions of face, body and hands as well as audio and multi-view raw video footage.
The limited ($\approx3$ hours) and specific data (only focused on haggling), makes it difficult to learn diverse motion distributions.

Inspired by this work, we introduce a medium-scale dataset capturing dyadic conversations between pairs of individuals totaling to 8 hours of video data from 4 participants, each engaging in 2 hours of paired conversational data. To ensure diversity of expressions and gestures, we prompt the actors with a diversity of situations such as selling, interviews, difficult scenarios, and everyday discourse. 

Most notably, to the best of our knowledge, we are the first to provide a dataset accompanied with fully photorealistic renderings of the conversational agents. 
Rather than generating and evaluating motion via 3D meshes, our multi-view dataset allows us to reconstruct the full face, body, and hands in a photorealistic manner. 
Visualizing via these renderings allow us to be more perceptive to fine-grain details in motion that are often missed when rendered via coarse 3D meshes.
Our evaluations confirm the importance of evaluating gestural motion using photo-real avatars. 

To create the photorealistic renderings, we captured both individuals simultaneously in multi-view capture domes. One person stood in a full-body dome while the other sat in a head-only dome. During the conversations, both viewed screens of the other person in real-time. 
We can then reconstruct high fidelity renderings of the face only for one individual~\cite{lombardi2018deep}, and the face, body, and hands for the other~\cite{bagautdinov2021bodies}. To train our method, we use the ground truth from the full-body capture to supervise our method.
We will publicly release audio, video, precomputed joint angles, face expression codes, and trained personalized avatar renderers.




\section{Experiments}
\label{sec:experiments}

We evaluate the ability of our model to effectively generate realistic conversational motion. 
We quantitatively measure the realism and diversity of our results against tracked ground truth data $(\mathbf{F}, \mathbf{J})$.
We also perform a perceptual evaluation to corroborate the quantitative results and to measure appropriateness of our generated gestures in the given conversational setting. 
Our results demonstrate evaluators are more perceptive to subtle gestures when rendered on photo-realistic avatars than on 3D meshes.

\subsection{Experimental Setup}
\paragraph{Evaluation Metrics.} Following a combination of prior works~\cite{ng2022learning, ao2023gesturediffuclip, alexanderson2023listen}, we use a composition of metrics to measure the realism and diversity of generated motion. 

\begin{itemize}[noitemsep,topsep=0pt,leftmargin=*]
    \item $\text{FD}_g$: ``geometric" realism measured by distribution distance between generated and ground truth \emph{static} poses. 
    We directly calculated the Frechet distance (FD) in the expression $\mathds{R}^{d_f}$ and pose space $\mathds{R}^{d_j \times 3}$.
    \item $\text{FD}_k$: ``kinetic" motion realism. Similar to above but distributions calculated on the velocities of motion sequences $\delta \mathbf{P}$. Computed in expression $\mathds{R}^{T \times d_f}$ and pose space $\mathds{R}^{T \times d_j \times 3}$. 
    \item $\text{Div}_g$: ``geometric" pose diversity. We randomly sample 30 expression, pose pairs within a motion sequence and compute average L2 distances between pairs to measure diversity of \emph{static} expressions/poses in the set.
    \item $\text{Div}_k$: Temporal variance across a sequence of expressions/poses. Measures amount of motion in a sequence. 
    \item $\text{Div}_\mathit{sample}$: Diversity across different samples. We group samples generated by the same audio and calculate variance across the samples.
\end{itemize}

Together, these metrics measure both the realism and diversity of the generated gestures in conversation.

\begin{table}\centering \footnotesize
\setlength{\tabcolsep}{5.0pt}
\ra{1.3} 
\begin{tabular}{lrrrrr}
\toprule
& $\text{FD}_g \downarrow$ & $\text{FD}_k \downarrow$ & $\text{Div}_g \uparrow$ & $\text{Div}_k \uparrow $ & $\text{Div}_\mathit{sample} \uparrow$\\
\cmidrule{2-6}
\rowcolor{Gray}
\textit{GT} & & & 3.09 & 2.50 &\\
Random & $9.37_{1.4}$ & $1.44_{0.04}$ & $3.10_{0.09}$ & $2.49_{0.4}$ & $3.97_{0.8}$ \\
KNN & $8.44_{1.6}$ & $0.62_{0.09}$ & $2.13_{0.05}$ & $1.21_{0.3}$ & $1.96_{0.3}$\\
SHOW~\cite{yi2023generating} & $4.97_{0.7}$ & $2.60_{0.10}$ & $2.10_{0.09}$ & $0.77_{0.1}$ & $2.82_{0.2}$\\
LDA~\cite{alexanderson2023listen} & $5.08_{0.2}$ & $1.04_{0.07}$ & $2.45_{0.06}$ & $1.88_{0.3}$ & $2.68_{0.4}$\\
\cmidrule{2-6}
Ours Uncond & $8.45_{1.3}$ & $1.53_{0.08}$ & $2.74_{0.07}$ & $2.06_{0.4}$ & $2.94_{0.3}$\\
Ours w/o $\mathbf{P}$  & $5.08_{0.4}$ & $1.13_{0.09}$ & $2.47_{0.06}$ & $1.67_{0.3}$ & $2.06_{0.4}$ \\
Ours w/o $\mathbf{A}$ & $3.94_{0.1}$ & $0.98_{0.10}$ & $2.69_{0.08}$ & $2.16_{0.4}$ & $2.71_{0.3}$ \\
\rowcolor{light-gray}
Ours & $2.94_{0.2}$ & $0.96_{0.07}$ & $2.98_{0.07}$ & ${2.36}_{0.4}$ & $3.58_{0.5}$\\
\bottomrule
\end{tabular}
\caption{\textbf{Baselines and ablations} vs.~ground truth poses (GT). $\downarrow$ indicates lower is better. We average across all subjects in the dataset. We sample 5 sequences for $\text{Div}_\text{sample}$ and average across all samples for each metric. Standard deviation as subscript $(\mu_{\sigma})$. 
}
\label{tab:results}
\vspace{-0.1cm}
\end{table}

\begin{table}\centering \footnotesize
\setlength{\tabcolsep}{5.0pt}.
\ra{1.3} 
\begin{tabular}{lrrr}
\toprule
&Horizontal L2 Error $\downarrow$ & Vertical L2 Error$\downarrow$ & Mesh~L2 $\downarrow$ \\
\cmidrule{2-4}
SHOW~\cite{yi2023generating} & 2.76 & 2.15 & 2.25 \\
Ours w/o $\mathbf{L}$ & 2.62 & 2.43 & 2.24 \\
\rowcolor{light-gray}
Ours & 2.29 & 1.89 & 1.76 \\
\bottomrule
\end{tabular}
\caption{\textbf{Lip reconstructions} 
The vertical (horizontal) distance is the distance between top and bottom (left and right) keypoints along the y (x) axis. The errors shown are L2 differences between ground truth and generated distances.
Mesh~L2 is the error in generated vs.~GT mesh vertices on the lip region. Errors in mm$^2$. 
}
\label{tab:lip}
\vspace{-0.25cm}
\end{table}

\paragraph{Baselines and ablations.} We compare to the following:

\begin{itemize}[noitemsep,topsep=0pt,leftmargin=*]
    \item \textbf{Random}: Random motion sequences from the train set.
    \item \textbf{KNN}: A segment-search method commonly used for synthesis. Given input audio, we find its nearest neighbor from the training set and use its corresponding motion segment as the prediction. We use audio features from Wav2Vec~\cite{baevski2020wav2vec} to encode the audio.
    \item \textbf{SHOW}~\cite{yi2023generating}: VQ-VAE based method that uses a transformer to autoregressively output motion conditioned on the audio of a speaker. They have separate models for face, body, and hands. Given~\cite{yi2023generating} is trained on monologues, we retrain their model for our domain.
    \item \textbf{LDA}~\cite{alexanderson2023listen}: Audio to motion diffusion model trained in a monologue setting. We re-train to adapt to our domain.
    \item \textbf{Ours Uncond}: (ablation) unconditional motion generation without audio or guide pose conditioning. 
    \item \textbf{Ours w/o $\mathbf{P}$}: (ablation) audio conditioned motion diffusion without guide pose conditioning. Similar to LDA~\cite{alexanderson2023listen}.
    \item \textbf{Ours w/o $\mathbf{A}$}: (ablation) guide pose conditioned motion diffusion model but without audio conditioning. Similar to diffusion infilling approaches.
\end{itemize}

\subsection{Results}
\label{sec:results}

Through quantitative evaluations, we show that our proposed method outputs realistic motion more diverse than competing baselines. In our Mechanical Turk A/B evaluations, we demonstrate our method generates compelling and plausible gestures, consistently outperforming our strongest baseline.
Additionally, the A/B tests highlight that photorealism effectively captures subtle nuances in gestures that are challenging to discern from 3D meshes. Yet these details significantly effect the evaluation of conversational motion.

\begin{figure}[t]
    \centering
    \includegraphics[width=\linewidth]{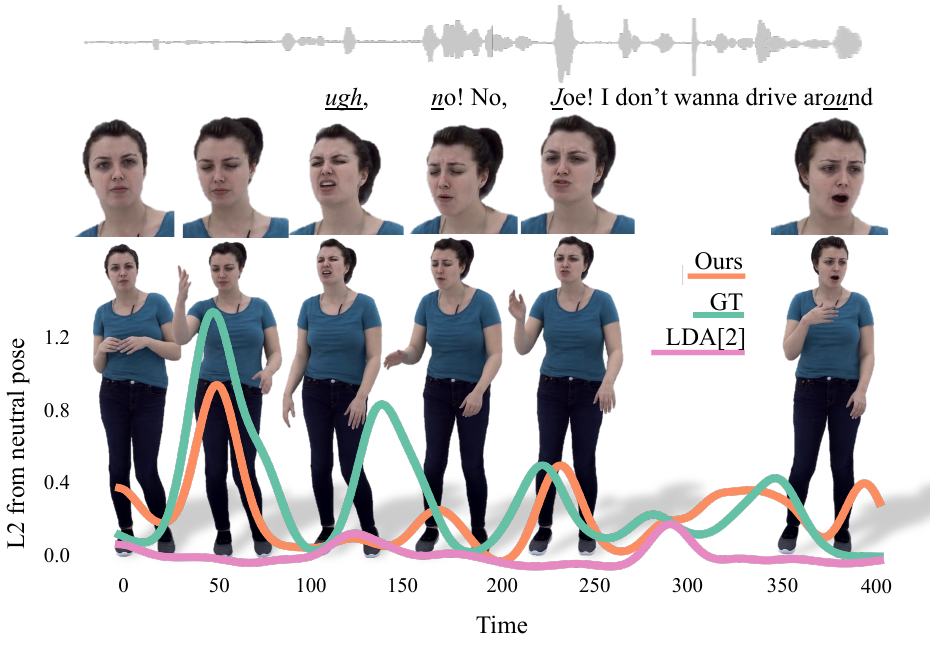}
    \vspace{-0.5cm}
    \caption{\textbf{Motion correlation with audio} Given audio (top), we plot the L2 distance of each pose to the mean neutral pose across 400 frames. Ours (rendered avatars, orange line) closely matches the peaks corresponding to large motion also seen in ground truth (e.g.~a flick of the hand preempting the ``ugh"). LDA~\cite{alexanderson2023listen} (pink) misses these peaky motions. Also note how our method generates highly expressive facial motion matching the speech. 
    }
    \label{fig:timeseries}
\end{figure}


\paragraph{Quantitative Results.}
\label{sec:quant}

Table~\ref{tab:results} shows that compared to prior works, our method achieves the lowest FD scores while generating motion with highest diversity.
While \textbf{Random} has good diversity that matches that of \textbf{GT}, 
the random segments do not appropriately match the corresponding conversational dynamics, resulting in 
higher $\text{FD}_g$. A slight improvement to \textbf{Random} is \textbf{KNN}, conventionally used for motion synthesis. While \textbf{KNN} performs better in terms of realism, matching the ``kinetic" distribution of the ground truth sequences better than \textbf{Ours}, the diversity across and within samples is significantly lower, also indicated by the higher ``geometric" FD. In Fig.~\ref{fig:keyframes}, we demonstrate the diversity of guide poses our method generates. Sampling via the VQ-based transformer $\mathcal{P}$ allows us to produce significantly different styles of poses conditioned on the same audio input. The diffusion model then learns to produce dynamic motion (Fig.~\ref{fig:analysis}), where the motion faithfully follows the conversational audio.

Our method outperforms both a VQ-only approach \textbf{SHOW}\cite{yi2023generating} and a diffusion-only approach \textbf{LDA}\cite{alexanderson2023listen}, achieving better realism and diversity across samples. Within sequences, our method generates more motion, resulting in a higher Div$_k$.
Fig.~\ref{fig:timeseries} highlights this, demonstrating \textbf{LDA}~\cite{alexanderson2023listen} produces dampened motion with less variation. In contrast, our method synthesizes variations in motion that closely match ground-truth.


\begin{figure}[t]
    \centering
    \includegraphics[width=\linewidth]{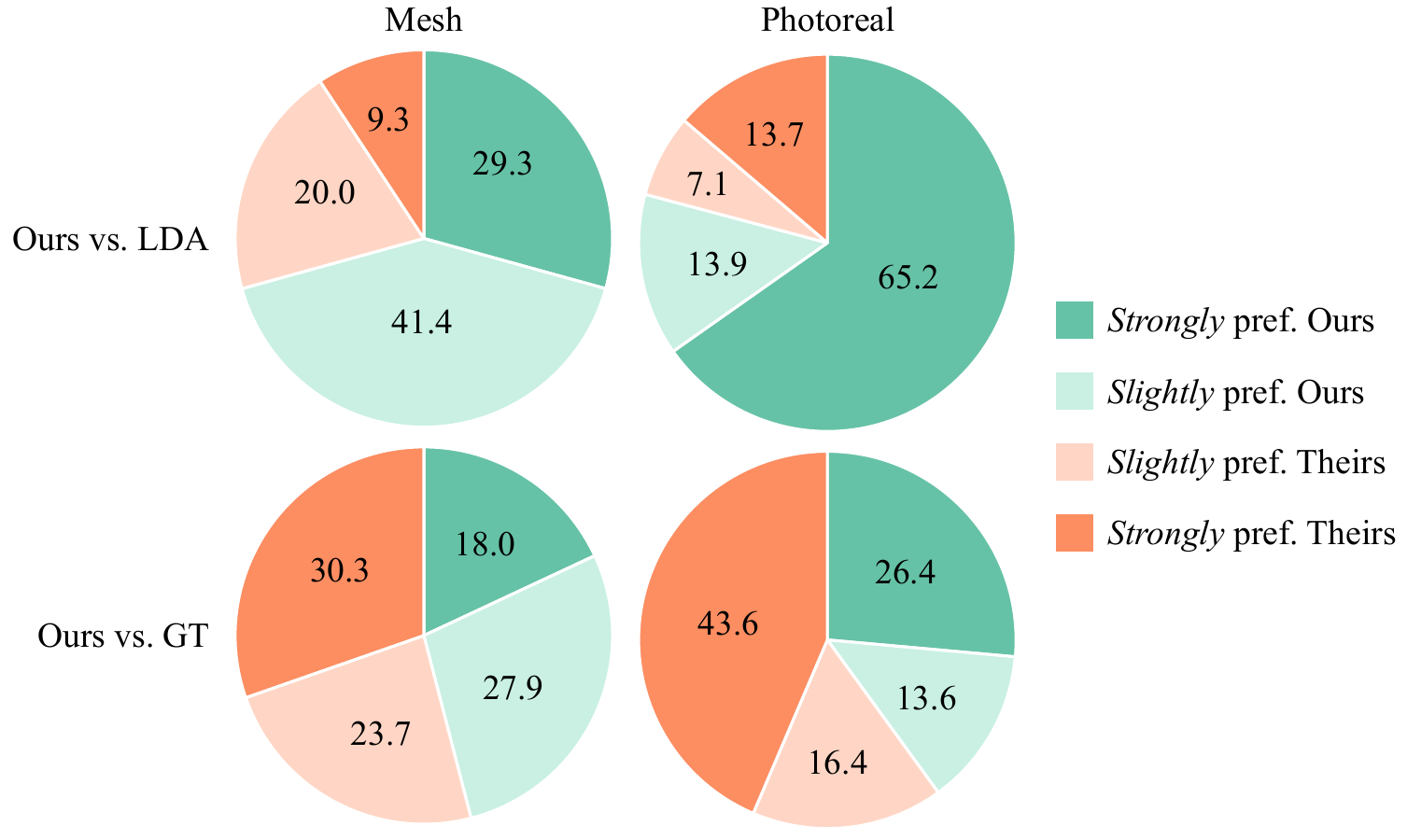}
    \caption{\textbf{Perceptual evaluation} on Ours vs.~ground truth or Ours vs.~our strongest baseline LDA~\cite{alexanderson2023listen}. We compare using mesh  vs.~photorealistic visualizations. Ours outperforms LDA~\cite{alexanderson2023listen} in both mesh and photoreal settings (top). Further, we note people are able to distinguish GT more often in the photoreal setting than with meshes (bottom). The results suggest that evaluating with photorealistic avatars leads to more accurate evaluations.
    }
    \label{fig:amt}
\end{figure}

Our ablations justify our design decisions. Applying our method without any conditioning (\textbf{Ours Uncond}) performs notably worse, with a realism and variance similar to that of \textbf{Random}. This suggests that while the motion generated does not match the given conversation sequence, it is similar to real motion in the dataset. Audio only conditioning (\textbf{Ours w/o $\mathbf{P}$}) improves over unconditional generation and performs similarly to \textbf{LDA}~\cite{alexanderson2023listen}, an audio to motion diffusion-based method. 
The lower diversity in both the static poses and across a temporal sequence results in higher FD scores.
When adding only the guide pose conditioning (\textbf{Ours w/o $\mathbf{A}$}), both the diversities and FD scores improve significantly. This suggests that the coarse-to-fine paradigm, introduced through the predicted guide poses, helps to add diversity to the diffusion results.
It also suggests that the coarse guide poses produced by the transformer $\mathcal{P}$ follow a trajectory that faithfully matches the dynamics of the conversational audio. The FD scores and diversities further improve when adding both the  audio and guide pose conditioning in the body motion model $\mathcal{J}$.

Furthermore, we analyze the accuracy of our method in generating lip motion. In Table~\ref{tab:lip}, we calculate the vertical and horizontal distances between two pairs of keypoints representing the top/bottom and left/right corners of the mouth, respectively. The vertical distance measures errors in mouth opening while the horizontal distance measures mouth expressions, \eg a smile shifts the positions of the left/right mouth corner and increases the horizontal distance. We compare these distances against ground truth and compute the L2 error. Our approach (\textbf{Ours} in Table~\ref{tab:lip}) substantially outperforms an ablation without the pretrained lip regressor (\textbf{Ours w/o $\mathbf{L}$} in Table~\ref{tab:lip}) and the baseline \textbf{SHOW}~\cite{yi2023generating}. Qualitatively, the pretraining of the lip regressor not only improves lip syncing, but also prevents the mouth from randomly opening and closing while not talking. This results in better overall lip reconstructions, with lower errors on the face mesh vertices (\textit{Mesh L2}).

\paragraph{Perceptual Evaluation.}
\label{sec:perceptual}

Given the challenge of quantifying the coherence of gestures in conversation, we primarily evaluate this aspect through a perceptual evaluation. We conducted two variations of A/B tests on Amazon Mechanical Turk. In the first, evaluators viewed motion rendered on a generic non-textured mesh. In the second, they viewed videos of the motion on photorealistic avatars. 

In both cases, evaluators watched a series of video pairs. In each pair, one video was from our model and the other was from either our strongest baseline LDA~\cite{alexanderson2023listen} or ground truth. 
Evaluators were then asked to identify the motion that looked more plausible given the conversational audio. They were also asked to indicate how confident they were in their answer by selecting ``slightly prefer" vs.~``strongly prefer". 

As shown in Fig.~\ref{fig:amt}, ours significantly outperforms against our strongest baseline LDA~\cite{yi2023generating}, with about $70\%$ of evaluators preferring our method in both the mesh and photoreal settings. Interestingly, evaluators shifted from \emph{slightly} to \emph{strongly} preferring ours when visualized in a photorealistic manner (top row). This trend continues when we compare our method against ground truth (bottom row). While ours performs competitively against ground truth in a mesh-based rendering, it lags in the photoreal domain with $43\%$ of evaluators \emph{strongly} preferring ground truth over ours. Since meshes often obscure subtle motion details, it is difficult to accurately evaluate the nuances in gestures leading to evaluators being more forgiving of ``incorrect" motions. Our results suggest that photorealism is essential to accurately evaluating conversational motion.

\section{Conclusion}
In this work, we explored generating conversational gestures conditioned on audio for fully embodied photorealistic avatars. 
To this end, we combine the benefits of vector quantization with diffusion to generate more expressive and diverse motion.
We train on a novel multi-view, long-form conversational dataset that allows for photorealistic reconstructions.
Our method produces diverse face, body, and hand motion that accurately matches the conversational dynamics. 
The results also underscore the significance of photorealism in evaluating fine-grain conversational motion.


\para{Limitations and ethical considerations.}
While our model produces realistic motion, it operates on short-range audio.
It thus fails to generate gestures requiring long-range language understanding (\eg~counting), which we leave for future work.
Further, our work is limited to photorealistic generation of four subjects in our dataset.
This limitation addresses ethical concerns since only consenting participants can be rendered, as opposed to arbitrary non-consenting humans.
In releasing a dataset with full participant consent, we hope to provide researchers the opportunity to explore photorealistic motion synthesis in an ethical setting.

\medskip
\noindent \textbf{Acknowledgements.} The work of Ng and Darrell is supported by DoD and/or BAIR Commons resources.

{\small
\bibliographystyle{ieee_fullname}
\bibliography{egbib}

\begin{thebibliography}{10}\itemsep=-1pt

\bibitem{ahuja2019react}
Chaitanya Ahuja, Shugao Ma, Louis-Philippe Morency, and Yaser Sheikh.
\newblock To react or not to react: End-to-end visual pose forecasting for personalized avatar during dyadic conversations.
\newblock In {\em 2019 International Conference on Multimodal Interaction}, pages 74--84, 2019.

\bibitem{alexanderson2023listen}
Simon Alexanderson, Rajmund Nagy, Jonas Beskow, and Gustav~Eje Henter.
\newblock Listen, denoise, action! audio-driven motion synthesis with diffusion models.
\newblock {\em ACM Transactions on Graphics (TOG)}, 42(4):1--20, 2023.

\bibitem{ao2023gesturediffuclip}
Tenglong Ao, Zeyi Zhang, and Libin Liu.
\newblock Gesturediffuclip: Gesture diffusion model with clip latents.
\newblock {\em arXiv preprint arXiv:2303.14613}, 2023.

\bibitem{baevski2020wav2vec}
Alexei Baevski, Yuhao Zhou, Abdelrahman Mohamed, and Michael Auli.
\newblock wav2vec 2.0: A framework for self-supervised learning of speech representations.
\newblock {\em Advances in neural information processing systems}, 33:12449--12460, 2020.

\bibitem{bagautdinov2021bodies}
Timur Bagautdinov, Chenglei Wu, Tomas Simon, Fabi\'{a}n Prada, Takaaki Shiratori, Shih-En Wei, Weipeng Xu, Yaser Sheikh, and Jason Saragih.
\newblock Driving-signal aware full-body avatars.
\newblock {\em ACM Trans. Graph.}, 40(4), 2021.

\bibitem{Bohus_Horvitz_2010}
Dan Bohus and Eric Horvitz.
\newblock Facilitating multiparty dialog with gaze, gesture, and speech.
\newblock In {\em International Conference on Multimodal Interfaces and the Workshop on Machine Learning for Multimodal Interaction}, ICMI-MLMI '10, New York, NY, USA, 2010. Association for Computing Machinery.

\bibitem{Cassell1994}
Justine Cassell, Catherine Pelachaud, Norman Badler, Mark Steedman, Brett Achorn, Tripp Becket, Brett Douville, Scott Prevost, and Matthew Stone.
\newblock Animated conversation: Rule-based generation of facial expression, gesture and spoken intonation for multiple conversational agents.
\newblock In {\em Proceedings of the 21st Annual Conference on Computer Graphics and Interactive Techniques}, SIGGRAPH '94, New York, NY, USA, 1994. Association for Computing Machinery.

\bibitem{chaminade2007anthropomorphism}
Thierry Chaminade, Jessica Hodgins, and Mitsuo Kawato.
\newblock Anthropomorphism influences perception of computer-animated characters’ actions.
\newblock {\em Social cognitive and affective neuroscience}, 2(3):206--216, 2007.

\bibitem{cudeiro2019voca}
Daniel Cudeiro, Timo Bolkart, Cassidy Laidlaw, Anurag Ranjan, and Michael Black.
\newblock Capture, learning, and synthesis of {3D} speaking styles.
\newblock In {\em Proceedings IEEE Conf. on Computer Vision and Pattern Recognition (CVPR)}, pages 10101--10111, 2019.

\bibitem{ekman1969nonverbal}
Paul Ekman and Wallace~V Friesen.
\newblock Nonverbal leakage and clues to deception.
\newblock {\em Psychiatry}, 32(1):88--106, 1969.

\bibitem{esser2021taming}
Patrick Esser, Robin Rombach, and Bjorn Ommer.
\newblock Taming transformers for high-resolution image synthesis.
\newblock In {\em Proceedings of the IEEE/CVF conference on computer vision and pattern recognition}, pages 12873--12883, 2021.

\bibitem{learn2smile2017}
Will Feng, Anitha Kannan, Georgia Gkioxari, and Larry Zitnick.
\newblock Learn2smile: Learning non-verbal interaction through observation.
\newblock {\em IROS}, 2017.

\bibitem{ginosar2019learning}
Shiry Ginosar, Amir Bar, Gefen Kohavi, Caroline Chan, Andrew Owens, and Jitendra Malik.
\newblock Learning individual styles of conversational gesture.
\newblock In {\em Proceedings of the IEEE/CVF Conference on Computer Vision and Pattern Recognition}, pages 3497--3506, 2019.

\bibitem{gratch2006virtual}
Jonathan Gratch, Anna Okhmatovskaia, Francois Lamothe, Stacy Marsella, Mathieu Morales, Rick~J van~der Werf, and Louis-Philippe Morency.
\newblock Virtual rapport.
\newblock In {\em International Workshop on Intelligent Virtual Agents}, pages 14--27. Springer, 2006.

\bibitem{ho2020denoising}
Jonathan Ho, Ajay Jain, and Pieter Abbeel.
\newblock Denoising diffusion probabilistic models.
\newblock {\em Advances in neural information processing systems}, 33:6840--6851, 2020.

\bibitem{ho2021classifier}
Jonathan Ho and Tim Salimans.
\newblock Classifier-free diffusion guidance.
\newblock {\em NeurIPS 2021 Workshop on Deep Generative Models and Downstream Applications}, 2021.

\bibitem{holtzman2019curious}
Ari Holtzman, Jan Buys, Li Du, Maxwell Forbes, and Yejin Choi.
\newblock The curious case of neural text degeneration.
\newblock {\em arXiv preprint arXiv:1904.09751}, 2019.

\bibitem{huang2011virtual}
Lixing Huang, Louis-Philippe Morency, and Jonathan Gratch.
\newblock Virtual rapport 2.0.
\newblock In {\em International workshop on intelligent virtual agents}, pages 68--79. Springer, 2011.

\bibitem{huang2017dyadgan}
Yuchi Huang and Saad~M Khan.
\newblock Dyadgan: Generating facial expressions in dyadic interactions.
\newblock In {\em Proceedings of the IEEE Conference on Computer Vision and Pattern Recognition Workshops}, pages 11--18, 2017.

\bibitem{ji2021audio}
Xinya Ji, Hang Zhou, Kaisiyuan Wang, Wayne Wu, Chen~Change Loy, Xun Cao, and Feng Xu.
\newblock Audio-driven emotional video portraits.
\newblock In {\em Proceedings of the IEEE/CVF conference on computer vision and pattern recognition}, pages 14080--14089, 2021.

\bibitem{jonell2019learning}
Patrik Jonell, Taras Kucherenko, Erik Ekstedt, and Jonas Beskow.
\newblock Learning non-verbal behavior for a social robot from youtube videos.
\newblock In {\em ICDL-EpiRob Workshop on Naturalistic Non-Verbal and Affective Human-Robot Interactions, Oslo, Norway, August 19, 2019}, 2019.

\bibitem{joo2019towards}
Hanbyul Joo, Tomas Simon, Mina Cikara, and Yaser Sheikh.
\newblock Towards social artificial intelligence: Nonverbal social signal prediction in a triadic interaction.
\newblock In {\em Proceedings of the IEEE/CVF Conference on Computer Vision and Pattern Recognition}, pages 10873--10883, 2019.

\bibitem{lee2019talking}
Gilwoo Lee, Zhiwei Deng, Shugao Ma, Takaaki Shiratori, Siddhartha~S Srinivasa, and Yaser Sheikh.
\newblock Talking with hands 16.2 m: A large-scale dataset of synchronized body-finger motion and audio for conversational motion analysis and synthesis.
\newblock In {\em Proceedings of the IEEE/CVF International Conference on Computer Vision}, pages 763--772, 2019.

\bibitem{liu2022beat}
Haiyang Liu, Zihao Zhu, Naoya Iwamoto, Yichen Peng, Zhengqing Li, You Zhou, Elif Bozkurt, and Bo Zheng.
\newblock Beat: A large-scale semantic and emotional multi-modal dataset for conversational gestures synthesis.
\newblock {\em arXiv preprint arXiv:2203.05297}, 2022.

\bibitem{lombardi2018deep}
Stephen Lombardi, Jason Saragih, Tomas Simon, and Yaser Sheikh.
\newblock Deep appearance models for face rendering.
\newblock {\em ACM Trans. on Graphics}, 37(4), 2018.

\bibitem{mcdonnell2012render}
Rachel McDonnell, Martin Breidt, and Heinrich~H B{\"u}lthoff.
\newblock Render me real? investigating the effect of render style on the perception of animated virtual humans.
\newblock {\em ACM Transactions on Graphics (TOG)}, 31(4):1--11, 2012.

\bibitem{ng2021body2hands}
Evonne Ng, Shiry Ginosar, Trevor Darrell, and Hanbyul Joo.
\newblock Body2hands: Learning to infer 3d hands from conversational gesture body dynamics.
\newblock In {\em Proceedings of the IEEE/CVF Conference on Computer Vision and Pattern Recognition}, pages 11865--11874, 2021.

\bibitem{ng2022learning}
Evonne Ng, Hanbyul Joo, Liwen Hu, Hao Li, Trevor Darrell, Angjoo Kanazawa, and Shiry Ginosar.
\newblock Learning to listen: Modeling non-deterministic dyadic facial motion.
\newblock In {\em Proceedings of the IEEE/CVF Conference on Computer Vision and Pattern Recognition}, pages 20395--20405, 2022.

\bibitem{Ng_2023_ICCV}
Evonne Ng, Sanjay Subramanian, Dan Klein, Angjoo Kanazawa, Trevor Darrell, and Shiry Ginosar.
\newblock Can language models learn to listen?
\newblock In {\em Proceedings of the IEEE/CVF International Conference on Computer Vision (ICCV)}, 2023.

\bibitem{nichol2021improved}
Alexander~Quinn Nichol and Prafulla Dhariwal.
\newblock Improved denoising diffusion probabilistic models.
\newblock In {\em International Conference on Machine Learning}, pages 8162--8171, 2021.

\bibitem{nojavanasghari2018interactive}
Behnaz Nojavanasghari, Yuchi Huang, and Saad Khan.
\newblock Interactive generative adversarial networks for facial expression generation in dyadic interactions.
\newblock {\em arXiv preprint arXiv:1801.09092}, 2018.

\bibitem{richard2021meshtalk}
Alexander Richard, Michael Zollhoefer, Yandong Wen, Fernando de~la Torre, and Yaser Sheikh.
\newblock Meshtalk: 3d face animation from speech using cross-modality disentanglement.
\newblock In {\em Proceedings of the IEEE/CVF International Conference on Computer Vision (ICCV)}, 2021.

\bibitem{tanke2023social}
Julian Tanke, Linguang Zhang, Amy Zhao, Chengcheng Tang, Yujun Cai, Lezi Wang, Po-Chen Wu, Juergen Gall, and Cem Keskin.
\newblock Social diffusion: Long-term multiple human motion anticipation.
\newblock In {\em Proceedings of the IEEE/CVF International Conference on Computer Vision}, pages 9601--9611, 2023.

\bibitem{tevet2022human}
Guy Tevet, Sigal Raab, Brian Gordon, Yonatan Shafir, Daniel Cohen-Or, and Amit~H Bermano.
\newblock Human motion diffusion model.
\newblock {\em arXiv preprint arXiv:2209.14916}, 2022.

\bibitem{thies2020neural}
Justus Thies, Mohamed Elgharib, Ayush Tewari, Christian Theobalt, and Matthias Nie{\ss}ner.
\newblock Neural voice puppetry: Audio-driven facial reenactment.
\newblock In {\em Computer Vision--ECCV 2020: 16th European Conference, Glasgow, UK, August 23--28, 2020, Proceedings, Part XVI 16}, pages 716--731. Springer, 2020.

\bibitem{tseng2023edge}
Jonathan Tseng, Rodrigo Castellon, and Karen Liu.
\newblock Edge: Editable dance generation from music.
\newblock In {\em Proceedings of the IEEE/CVF Conference on Computer Vision and Pattern Recognition}, pages 448--458, 2023.

\bibitem{van2017neural}
Aaron Van Den~Oord, Oriol Vinyals, et~al.
\newblock Neural discrete representation learning.
\newblock {\em Advances in neural information processing systems}, 30, 2017.

\bibitem{vasuki2006review}
A Vasuki and PT Vanathi.
\newblock A review of vector quantization techniques.
\newblock {\em IEEE Potentials}, 25(4):39--47, 2006.

\bibitem{vougioukas2020realistic}
Konstantinos Vougioukas, Stavros Petridis, and Maja Pantic.
\newblock Realistic speech-driven facial animation with gans.
\newblock {\em International Journal of Computer Vision}, 128:1398--1413, 2020.

\bibitem{yi2023generating}
Hongwei Yi, Hualin Liang, Yifei Liu, Qiong Cao, Yandong Wen, Timo Bolkart, Dacheng Tao, and Michael~J Black.
\newblock Generating holistic 3d human motion from speech.
\newblock In {\em Proceedings of the IEEE/CVF Conference on Computer Vision and Pattern Recognition}, pages 469--480, 2023.

\bibitem{yu2023talking}
Zhentao Yu, Zixin Yin, Deyu Zhou, Duomin Wang, Finn Wong, and Baoyuan Wang.
\newblock Talking head generation with probabilistic audio-to-visual diffusion priors.
\newblock In {\em Proceedings of the IEEE/CVF International Conference on Computer Vision}, pages 7645--7655, 2023.

\bibitem{zeghidour2021soundstream}
Neil Zeghidour, Alejandro Luebs, Ahmed Omran, Jan Skoglund, and Marco Tagliasacchi.
\newblock Soundstream: An end-to-end neural audio codec.
\newblock {\em IEEE/ACM Transactions on Audio, Speech, and Language Processing}, 30:495--507, 2021.

\bibitem{zhang2020apb2face}
Jiangning Zhang, Liang Liu, Zhucun Xue, and Yong Liu.
\newblock Apb2face: Audio-guided face reenactment with auxiliary pose and blink signals.
\newblock In {\em ICASSP 2020-2020 IEEE International Conference on Acoustics, Speech and Signal Processing (ICASSP)}, pages 4402--4406. IEEE, 2020.

\bibitem{Zhi_2023_ICCV}
Yihao Zhi, Xiaodong Cun, Xuelin Chen, Xi Shen, Wen Guo, Shaoli Huang, and Shenghua Gao.
\newblock Livelyspeaker: Towards semantic-aware co-speech gesture generation.
\newblock In {\em Proceedings of the IEEE/CVF International Conference on Computer Vision (ICCV)}, pages 20807--20817, October 2023.

\bibitem{zhou2022responsive}
Mohan Zhou, Yalong Bai, Wei Zhang, Ting Yao, Tiejun Zhao, and Tao Mei.
\newblock Responsive listening head generation: A benchmark dataset and baseline.
\newblock In {\em ECCV}, 2022.

\end{thebibliography}
}

\clearpage

{\Large{\textbf{Appendix}}}
\appendix
\section{Results 
Video} 
\vspace{-0.2cm}
\label{app:vid}
The supplementary video shows sequences of various individuals in different conversational settings from our dataset. Below, we denote the time stamp range associated with the discussion - (@mm:ss-mm:ss). 

The results show that our model successfully models plausible, motion that is synchronous with the ongoing conversational dynamics. For instance, it correctly generates facial expressions and body language of someone feeling disgruntled~\eg dismissive hand wave and turning away (@02:58-03:11). The generated gestures are well-timed with the conversation~\eg raised finger with ``I think" (@02:30-@02:50). Additionally, our approach can produce multiple plausible motion trajectories based on a single conversational audio input, each with distinct variations (@03:15-@03:55). 

Compared against baselines and prior works, our method generates more ``peaky" motion such as wrist flicks while listing (@04:16), and finger pointing (@04:50), which are both missed by a diffusion-based method LDA~[Alexanderson~\etal 2023]. In comparison to a VQ-based method SHOW~[Yi~\etal 2023], ours produces more dynamic motion with increased arm movement (@04:52), and seamless transitions between poses when switching from asking a question, to listening, to responding (@05:12-05:30). In contrast, SHOW moves to the audio but hovers around the same pose throughout. In comparison to both Random and KNN, gestures by our approach match the audio far better.

Notably, without any retraining, our method generalizes to conversational audio not seen in the dataset, such as a random movie clip audio (@05:44-@06:03). This is possibly due to the identity-agnostic training of Wav2Vec.
We can also extend our method to the application of video editing, where we can reanimate a target person with a different motion trajectory by swapping guide poses (@06:10-06:27). 

\section{Method}
\label{app:method}
\vspace{-0.2cm}
\subsection{Pose representation}
\label{app:pose}
While we use a standard SO(3) representation for the joint angles, we note that not all joints are parameterized with 3 degrees of freedom (~\eg arm twist is only represented with roll, head bend with yaw, etc.~). In total, we have 104 rotation angles across all of the joints.

\subsection{Residual VQ-VAE}
\label{app:vq}
The residual VQ-VAE allows us to capture finer-grain details by employing a cascade of codebooks to capture progressively finer approximations.
We use residual length of 4. In practice, this means we need a sequence of 4 VQ tokens to represent a single pose.
To generate poses during test time for the diffusion model, we autoregressively output $4 \times K$ tokens one at a time, where $K$ is the length of the downsampled sequence.
For the both the encoder and decoder, we use a series of 1D convolutions of kernel size 2. The total receptive field for both the encoder and decoder is 8. We use a codebook size of 1024, and embedding size of 64. We train for 300k steps.

\subsection{Guide pose Transformer}
\label{app:guide}
We adapt the diffusion model's architecture for the guide pose network. The transformer architecture is composed of masked self-attention layers that focuses only on previous timesteps to enable autoregressive prediction. The audio is then incorporated using non-causal cross attention layers. This means the network doesn't see past motion, but sees the full context of audio. We then remove the diffusion timestep $\tau$ conditioning, and instead feed in an audio embedding (averaged over the whole time series) to the FiLM layers. While not necessary, this slightly helps the transformer to generate more plausible poses on the very first time-step. We use 2 masked self-attention layers and 6 cross-attention layers, all with 8 heads. 
We train for $~\approx 100k$ iterations depending on the individual.

\subsection{Implementation details}
We use a max sequence length of 600 frames at 30 fps (20 second videos). 
During training, we randomly sample a sequence between 240 frames and 600 frames.
We then train on padded sequences of random lengths for all of our networks.
This allows us to generate sequences of arbitrary length during test time.
We train each network for each subject in the data separately.
All networks are trained on a single A100. Approximate train times: face diffusion model (8 hr), VQ + coarse pose predictor (5 hr), pose diffusion model (8 hr).

\section{Results}
\vspace{-0.2cm}
\label{app:results}

\subsection{Perceptual evaluation}
\label{app:amt}
For each Ours vs.~GT (mesh), vs.~GT (photoreal), vs.~LDA (mesh), vs.~LDA (photoreal), we generate 50 A-B tests. For each test, we ask 3 different evaluators, totalling to 600 evaluators. Each A-B test contained 14 questions.
Prior to the actual test, we provide a headphone check to make sure the evaluators are listening to audio. However, we do not ask additional questions that check to see if they are actually listening to the speech. The landing page describes the task and walks evaluators through 2 examples. 
To ensure the evaluators are not randomly clicking, we include 3 questions with an obvious mismatch (one speaker laughing while the listener is neutral) twice. If the evaluator selects a different response for these duplicated questions, we do not allow them to submit.

\subsection{Ablation with VQ-only method}
\label{app:ablation}
In the main paper, the VQ-only baseline is represented with prior work SHOW [Alexanderson~\etal 2023], which is very similar to our guide pose network. For completeness, we also train a VQ-only baseline using our network architecture. We see very similar results to SHOW and similar limitations. Quantitatively, FD$_g=5.00$, FD$_k=2.80$, Div$_g=2.20$, Div$_k=1.89$. Note the higher FD and lower diversity compared to our complete method. We notice that after many timesteps, drift often happens which causes the method to either get stuck in a local minima (no motion). 


\end{document}